\documentclass[10pt, a4paper]{article}

\newcommand\blfootnote[1]{%
  \begingroup
  \renewcommand\thefootnote{}\footnote{#1}%
  \addtocounter{footnote}{-1}%
  \endgroup
}

\usepackage[table,xcdraw]{xcolor}
\usepackage{amsmath}
\usepackage{multirow}

\usepackage{lrec-coling2024}

\title{A Bit of a Problem: Measurement Disparities in Dataset Sizes Across Languages}

\name{Catherine Arnett*$^1$, Tyler A. Chang*$^{2,3}$, Benjamin K. Bergen$^2$}

\address{$^1$ Department of Linguistics, \\
        $^2$ Department of Cognitive Science, \\
        $^3$ Halıcıoğlu Data Science Institute \\
        UC San Diego \\
         \{ccarnett, tachang, bkbergen\}@ucsd.edu\\}

\abstract{How should text dataset sizes be compared across languages?
Even for content-matched (parallel) corpora, UTF-8 encoded text can require a dramatically different number of bytes for different languages.
In our work, we define the byte premium between two languages as the ratio of bytes used to encode content-matched text in those languages.
We compute byte premiums for 1155 languages, and we use linear regressions to estimate byte premiums for other languages.
We release a tool to obtain byte premiums for any two languages, enabling comparisons of dataset sizes across languages for more equitable multilingual model development and data practices.
\\
\newline  }

\begin{document}

\maketitleabstract

\section{Introduction} \label{intro}
Large\blfootnote{*Equal contribution.} language datasets serve as the foundation for modern natural language technologies.
However, an often ignored question is how to compare dataset sizes across languages.
For standard multilingual language models such as XLM-R, BLOOM, and XGLM, dataset sizes are reported in bytes (\citealp{conneau-etal-2020-unsupervised,workshop2022bloom,lin-etal-2022-shot}).\footnote{Dataset sizes are also often reported in tokens, which depend on model-specific tokenizers and which exhibit similar cross-language disparities to bytes \citep{petrov2024language}.}
However, content-matched (i.e. parallel) text in two languages does not generally have the same size in bytes, with some languages taking over $5\times$ as many bytes as others (\S\ref{sec:computing-byte-premiums}).

Here, we compute \textbf{byte premiums} (cf. tokenization premiums in \citealp{petrov2024language}), the ratios of bytes taken to encode text in 1155 different languages.
We find that these byte premiums are highly correlated across datasets.
We fit linear regressions to estimate byte premiums for languages not included in our parallel datasets, and we release a simple Python tool to retrieve or predict the byte premium between any two languages.\footnote{ \url{https://github.com/catherinearnett/byte-premium-tool}}
Our work enables comparisons of dataset sizes across languages, with implications for equitable multilingual model development and resource distribution.

\section{Related Work}
Using UTF-8 encoding, which is by far the most widespread text encoding \citep{unicode-davis-2012}, characters take between one and four bytes to encode \citep{unicode-consortium}. Numbers and Latin characters without diacritics are one byte, and all non-Latin scripts use two or more bytes per character.
This alone introduces a disparity in measured dataset sizes in bytes \citep{costa-jussa-etal-2017-byte}, but it must be balanced with the fact that different scripts encode different amounts of ``information'' per character.
For example, Mandarin has high UTF-8 bytes-per-character, but it generally requires fewer characters than Latin-script languages to encode the same content.
To account for this tradeoff, previous work has used parallel text, finding that byte-level tokenizers encode parallel text in some languages using more ``tokens'' (bytes) than others (``tokenization premiums''; \citealp{petrov2024language}).
We tie these results to dataset storage and training dataset size measurement, we compute the byte premium for 1155 languages, and we present a method to predict the byte premium for novel languages.
All our results use UTF-8 encoded text.

\section{Computing Byte Premiums} \label{sec:computing-byte-premiums}

In this section, we calculate the \textbf{byte premium} $\textrm{BP}_{A/B}$ for different language pairs, which we define as the ratio of bytes taken to encode a comparable amount of information in language $A$ relative to language $B$.
For example, if $A$ on average takes twice as many UTF-8 bytes to encode the same information (parallel text) as $B$, then $\textrm{BP}_{A/B}$ would be $2.0$.
These byte premiums are useful when measuring ``how much'' content is in each language in a corpus.
In multi-parallel corpora, we note that the byte premiums must satisfy:
\begin{equation}
\label{eq:pairwise-bp}
\textrm{BP}_{A/B} = 
% \frac{\textrm{Bytes}_{A}}{\textrm{Bytes}_{B}} = 
\frac{\textrm{Bytes}_{A}}{\textrm{Bytes}_{C}} * \frac{\textrm{Bytes}_{C}}{\textrm{Bytes}_{B}} = \frac{\textrm{BP}_{A/C}}{\textrm{BP}_{B/C}}
\end{equation}
This implies that if the byte premium is known for every language relative to some language $C$, then all pairwise byte premiums are determined.
Thus, we only calculate a single byte premium $\textbf{\textrm{BP}}_A = \textrm{BP}_{A/C}$ per language, all relative to reference language $C$.
We use $C=$ English as our reference language, but using any other reference language $C_0$ would simply multiply all our byte premiums by a constant $\textrm{BP}_{C/C_0}$.
In later sections, we refer to byte premiums relative to English unless otherwise noted.
In contrast to \citet{petrov2024language}, calculating a single byte premium per language allows byte premiums to be used for multilingual corpora beyond just pairwise corpora.\footnote{For example, if Equation \ref{eq:pairwise-bp} does not hold, then English-Mandarin and Arabic-Mandarin byte premiums could produce conflicting comparable dataset sizes when adding Mandarin data to an English+Arabic corpus.}

\subsection{NLLB}
\label{sec:nllb}
Computing byte premiums requires parallel corpora in the desired languages.
We first use NLLB \cite{costa2022no}, a dataset of pairwise parallel text segments in 188 languages.
We sample the first 100K parallel text segments for each language pair $(A, B)$, and we compute $\textrm{BP}_{A/B}$ as the mean ratio of bytes used in language $A$ versus $B$, averaged over all segments.
This produces a byte premium value for every language pair.

To fit a single byte premium $\textrm{BP}_A = \textrm{BP}_{A/C}$ for each language relative to a reference language $C$ (in our case English), we minimize the mean squared error of $\textrm{BP}_A / \textrm{BP}_B$ relative to the ground truth $\textrm{BP}_{A/B}$ (Equation \ref{eq:pairwise-bp}) over all language pairs $(A, B)$.
In other words, we fit 188 byte premium values (one per language) based on all 2656 pairwise byte premium values.
Fitting these single byte premiums ensures that Equation \ref{eq:pairwise-bp} holds for all pairs.

Byte premiums computed from NLLB are reported in Appendix Table \ref{tab:nllb_byte_premiums}.
For example, Burmese has byte premium $5.10$, so on average it takes $5.10\times$ as many UTF-8 bytes to encode text in Burmese versus English.
These byte premiums are consistent when computed from different subsets of the NLLB corpus, with Pearson's $r>0.999$ for byte premiums computed from ten disjoint subsets of 10\% of the NLLB corpus.
Notably, byte premiums computed from only 100 lines of text per language pair correlate with the byte premiums computed from the full NLLB dataset with Pearson's $r=0.955$, indicating that byte premiums can be computed from fairly small parallel corpora.

\subsection{Other Parallel Corpora} \label{sec:other-parallel-corpora}

For comparison, we also compute byte premiums from three multi-parallel corpora: FLORES-200 (\citealp{costa2022no}; 204 languages), the Bible (\citealp{eBible}; 1027 languages), and the Universal Declaration of Human Rights (\citealp{Vatanen10LREC}; UDHR; 241 languages).
For each language $A$ in each dataset, we compute $\textrm{BP}_A = \textrm{Bytes}_A / \textrm{Bytes}_C$ relative to reference language $C=$ English.
% Unlike the pairwise NLLB dataset,
Because each dataset is comprised of parallel text across all included languages, these byte premiums already satisfy Equation \ref{eq:pairwise-bp}.

Computed byte premiums are highly correlated between NLLB, FLORES, and the Bible (Table \ref{tab:bp_correlations}; Pearson's $r>0.90$), suggesting that byte premiums are fairly consistent across datasets.
We posit that lower correlations with UDHR byte premiums may be because the UDHR corpora are much shorter (roughly twenty total lines of text) and potentially more domain-specific than the other corpora. For this reason, we do not use UDHR in later sections.

\setlength{\belowcaptionskip}{-0.3cm}
\begin{table}[t]
    \centering
    \footnotesize
    \renewcommand{\arraystretch}{1.3}
    \begin{tabular}{|p{1.9cm}|cccc|}
        \cline{2-5}
        \multicolumn{1}{c|}{} &
        \multicolumn{1}{c|}{\rotatebox{90}{ NLLB }} &
        \multicolumn{1}{c|}{\rotatebox{90}{ FLORES }} &
        \multicolumn{1}{c|}{\rotatebox{90}{ Bible }} &
        \multicolumn{1}{c|}{\rotatebox{90}{ UDHR }} \\
        \hline
        FLORES & \textbf{0.919} & & \textbf{0.938} & 0.737 \\
        Bible & \textbf{0.921} & \textbf{0.938} & & 0.177 \\
        UDHR & 0.592 & 0.737 & 0.177 &  \\
        \hline
    \end{tabular}
    \normalsize
    \caption{Pearson correlations between byte premiums calculated from different datasets. Correlations are high between NLLB, FLORES, and the Bible.}
    \label{tab:bp_correlations}
\end{table}
\setlength{\belowcaptionskip}{-0.0cm}

\subsection{Byte Premiums After Compression}
Interestingly, we find that byte premiums persist after compression with the common compression algorithm $\texttt{gzip}$ (at maximum compression level 9).
When byte premiums are computed from the compressed FLORES corpora, they correlate strongly with the uncompressed byte premiums (Pearson's $r=0.890$).
However, the scale of variation across languages reduces substantially after compression; for example, uncompressed byte premiums of $4.0$ are roughly analogous to compressed byte premiums of $1.7$ (i.e. compressed data in that language takes only $1.7 \times$ as many bytes as the reference language rather than $4.0 \times$ as many bytes; Appendix \ref{app:compression}).
This suggests that standard compression algorithms reduce but do not fully alleviate disparities in dataset storage sizes across languages.

\section{Predicting Novel Byte Premiums} \label{sec:estimating}
In many cases, we may want to compute the byte premium for a language $A$ outside of our existing datasets.
If a single parallel text is available from $A$ to any language $B$ in our datasets, then the byte premium can easily be calculated as (using reference language $C$ as before):
\begin{equation}
\textrm{BP}_A= \frac{\textrm{Bytes}_A}{\textrm{Bytes}_C} = \frac{\textrm{Bytes}_A}{\textrm{Bytes}_B} * \textrm{BP}_B
\end{equation}
However, there may be cases where no parallel text is available for language $A$.
In this scenario, we can break the byte premium into (1) the mean bytes-per-character in $A$ and $C$, and (2) the ratio of characters needed to express the same information in $A$ and $C$ (the ``length ratio''):
\begin{equation}
\label{eq:breakdown-bp}
\textrm{BP}_A = \frac{\textrm{Bytes}_A}{\textrm{Bytes}_C} = \frac{\textrm{Bytes}_A}{\textrm{Chars}_A} * \frac{\textrm{Chars}_A}{\textrm{Chars}_C} * \frac{\textrm{Chars}_C}{\textrm{Bytes}_C}
\end{equation}
\setlength{\belowdisplayskip}{0cm}
\setlength{\abovedisplayskip}{0cm}
The bytes-per-character ratio for $A$ can be calculated with only monolingual text in $A$.
We find that this ratio is highly consistent regardless of the dataset used.
The computed bytes-per-character ratios correlate strongly (Pearson's $r>0.99$) when calculated from any of NLLB, the Bible, or FLORES with 20, 200, or 2000 lines of text.
Given the consistency of these bytes-per-character ratios, we find it efficient to break byte premiums down as in Equation \ref{eq:breakdown-bp} such that we only need to predict the length ratio between languages.

\subsection{Predicting Length Ratios}

We use linear regressions including language family, script (writing system), script type (e.g. alphabet vs. logography), and entropy over characters to predict the length ratio $\textrm{Chars}_A / \textrm{Chars}_C$ for a language $A$ relative to the reference language $C=$ English.
From the predicted length ratio, we can use Equation \ref{eq:breakdown-bp} to calculate the predicted byte premium for language $A$.
Our results use length ratios, bytes-per-character ratios, and character entropies computed from NLLB, FLORES, or the Bible when available, in order of decreasing priority.\footnote{As with byte premiums, the choice of reference language $C$ only multiplies all length ratios by a constant.
NLLB length ratios are computed in the same way as byte premiums, but using characters instead of bytes.
We obtain similar regression results using length ratios, bytes-per-character ratios, and character entropies computed from NLLB, FLORES, or the Bible (Appendix \ref{app:val_diff_datasets}).}

\paragraph{Language Family}
We predict that typological features (e.g. inflection patterns or morpho-syntactic distinctions) may drive differences in length ratios. Languages that are in the same language family are more likely to share typological features due to their shared historical origin \citep{moravcsik2012introducing}. 

\paragraph{Script and Script Type}
Some writing systems may encode higher information content per character than others (e.g. Chinese characters; \citealp{perfetti2005orthography}), which leads to low length ratios, because the same content takes fewer characters to write.
We separate scripts into four script types (alphabet, abjad, abugida, and logography; Appendix \ref{app:writing_systems}), and we use script type as a predictor for length ratio.
We also consider the specific script as a nested predictor (e.g. Latin vs. Cyrillic).

\paragraph{Character Entropy}
It has been proposed that languages with fewer phonemes (contrastive sounds) in their inventories have longer words, because it requires more sounds per word to generate the number of contrastive sound sequences necessary to communicate \citep{nettle1995segmental}.\footnote{We also measure the number of phonemes per language (PHOIBLE; \citealp{moran2014phoible}), but it does not help predict length ratios ($R^2 = 0.002$). Therefore we do not include it in our linear regressions.}
Using the same logic, we predict that a language that tends to use fewer unique characters will require longer character sequences to express information (a high length ratio).
We operationalize the number of unique characters in a language as the entropy over the character probability distribution in that language.
A higher entropy indicates that the probability distribution is spread across more characters.
Similar to bytes-per-character ratios (\S\ref{sec:estimating}), the entropy over characters is highly stable across datasets, even computed from as few as 20 lines of text (Pearson's $r > 0.90$ for the same datasets as \S\ref{sec:estimating}). 

\bigskip

We fit linear regressions to predict length ratios from three different subsets of our predictors.
This allows us to predict novel byte premiums depending on the available information about the novel language.
We consider the following three subsets: (I) character entropy, language family, script, and script type, (II) character entropy, script, and script type, and (III) character entropy and script type.
The predicted length ratios can be used to predict byte premiums using Equation \ref{eq:breakdown-bp}.

\section{Evaluating Byte Premium Predictions}
\label{sec:evaluation}
We validate the byte premium predictions from our linear regressions by looping through languages with known byte premiums (from NLLB, FLORES, or the Bible, in that order of priority), evaluating the byte premium prediction for that language when holding it out from regression fitting.\footnote{To prevent skew of regression coefficients, we clip byte premiums to a maximum of $4.0$ (three languages; Appendix \ref{app:nllb_byte_premium_table}).}
We report the root mean squared error (RMSE) for the three linear regressions described in the previous section (I, II, and III).
We compute separate RMSEs for (1) languages whose script is shared by less than five languages in our datasets, and (2) languages whose script is shared by five or more languages in our datasets.
Languages whose script is uncommon may have more poorly fitted script coefficients (and potentially language family coefficients), so we might expect them to exhibit larger byte premium prediction errors.

\setlength{\belowcaptionskip}{-0.4cm}
\begin{table}[t]
    \centering
    \footnotesize
    \renewcommand{\arraystretch}{1.3}
    \begin{tabular}{|p{3.0cm}|ccc|}
    \cline{2-4}
    \multicolumn{1}{c|}{} & \multicolumn{3}{c|}{Regression} \\
        \multicolumn{1}{c|}{} &
        \multicolumn{1}{c}{I} &
        \multicolumn{1}{c}{II} &
        \multicolumn{1}{c|}{III} \\
        \hline
        Scripts with count $\geq5$ & \textbf{0.261} & 0.288 & 0.290 \\
        \hline
        Scripts with count $<5$ & 0.770 & 0.739 & \textbf{0.589} \\
        \hline
    \end{tabular}
    \normalsize
    \caption{RMSEs when predicting byte premiums using different regressions, for languages with common and uncommon scripts.}
    \label{tab:comparing_datasets}
\end{table}
\setlength{\belowcaptionskip}{-0.0cm}

Results are reported in Table \ref{tab:comparing_datasets}.
For languages with common scripts (scripts with count $\geq5$), the regressions improve as predictors are added (III, II, then I).
For these languages, RMSEs reach $0.261$, indicating that the predicted byte premiums are on average approximately $0.261$ away from the true byte premiums.

As expected, we also find that languages with uncommon scripts (scripts with count $<5$) have higher errors in their predicted byte premiums, indicating that their script and family coefficients are poorly fitted.
For those languages, the regression with the lowest validation error is regression III, which only includes character entropy and script type as predictors.
The validation RMSE is $0.589$, indicating that predicted byte premiums for languages with uncommon scripts are on average approximately $0.589$ away from the true byte premiums.
Given that byte premiums can range from below $0.75$ to over $5.00$, even this simple regression is a substantial improvement over a naive assumption that languages take equal bytes to encode information (i.e. byte premium $1.0$). 
 
\section{Introducing the Tool}
Finally, we introduce a Python tool that returns pre-computed or predicted byte premiums for any language pair. The tool is available at \textcolor{red}{\url{https://github.com/catherinearnett/byte-premium-tool}}.
If both input languages are in our set of 1155 languages, the pairwise byte premium is computed from Equation \ref{eq:pairwise-bp} using our pre-computed byte premiums.
If exactly one language in the pair is not in our set of 1155 languages, the byte premium is computed from a user-provided parallel text (if available) from the novel language to any of our 1155 languages.
If no parallel text is available, the tool asks for a small monolingual corpus in the novel language(s), from which it can compute the character entropy and bytes-per-character ratio per language, to use in the regressions from \S\ref{sec:estimating}.
Following the validation results in \S\ref{sec:evaluation}, the tool uses regression I, II, or III (in order of decreasing priority) for languages with common scripts.
For languages with uncommon scripts, regression III is always used.
Aside from character entropy (which is computed from the user-provided monolingual text), regression III requires only the script type for the novel language(s), which can easily be found on sites such as Wikipedia.
Thus, our tool provides a simple interface from which to obtain the pairwise byte premium between any two languages, enabling easy dataset size conversions.

\section{Discussion and Conclusion}
\paragraph{Measuring Dataset Sizes} One implication of our work is that researchers currently may overestimate the amount of data that multilingual NLP models are trained on for non-Latin script languages (languages with high byte premiums).
These languages are often already underrepresented in NLP \citep{van-esch-etal-2022-writing}.
For example, if it is reported that a model is trained on 1GB of Georgian data, then based on its byte premium of $4.34$ relative to English, we should consider the model to be effectively trained on the Georgian equivalent of about 230MB of English data.

As a preliminary investigation into whether scaling training data quantities by byte premiums per language is indeed a ``better'' measure of training data quantity, we use this scaled measure to predict multilingual language model performance on various per-language benchmarks.
Across models and tasks, we find that the scaled data proportions do predict performance in different languages better than reported proportions, but not significantly ($p=0.13$; see Appendix \ref{app:downstream_performance} for details). 

\paragraph{Byte-Level Tokenization} Our results also have implications for dataset tokenization.
Previous work has argued that byte-level tokenizers enable more uniform treatment of different languages in a model \citep{zhang-xu-2022-byte,xue-etal-2022-byt5}, but our byte premiums demonstrate that some languages may still be at a disadvantage with byte-level tokenizers.
Tokenization length inequalities can lead to higher costs, longer latencies, and restricted effective context lengths for some languages \citep{ahia2023all,petrov2024language}, in this case languages with high byte premiums.

\paragraph{Equitable Resource Costs}
Finally, languages with high byte premiums require more storage space than other languages to store comparable content, and they are likely to require higher bandwidth connections to transmit text content.
In cases where storage is charged per (giga)byte or Internet connections are charged based on bandwidth and usage, uniform pricing rates across languages may lead to higher technology costs for low-resource language communities.
While only a marginal step towards solving such issues, our work makes it possible to take byte premiums into account when measuring text data sizes across languages.

\section{Acknowledgements}
We would like to thank the other members of the UCSD Language and Cognition Lab for valuable discussion.
Tyler Chang is partially supported by the UCSD Halıcıoğlu Data Science Institute graduate fellowship.

\section{Bibliographical References} \label{sec:reference}

\bibliographystyle{lrec-coling2024-natbib}
\bibliography{custom}

\begin{thebibliography}{26}
\expandafter\ifx\csname natexlab\endcsname\relax\def\natexlab#1{#1}\fi

\bibitem[{Ahia et~al.(2023)Ahia, Kumar, Gonen, Kasai, Mortensen, Smith, and Tsvetkov}]{ahia2023all}
Orevaoghene Ahia, Sachin Kumar, Hila Gonen, Jungo Kasai, David Mortensen, Noah Smith, and Yulia Tsvetkov. 2023.
\newblock \href {https://doi.org/10.18653/v1/2023.emnlp-main.614} {Do all languages cost the same? {T}okenization in the era of commercial language models}.
\newblock In \emph{Proceedings of the 2023 Conference on Empirical Methods in Natural Language Processing}, pages 9904--9923, Singapore. Association for Computational Linguistics.

\bibitem[{Conneau et~al.(2020)Conneau, Khandelwal, Goyal, Chaudhary, Wenzek, Guzm{\'a}n, Grave, Ott, Zettlemoyer, and Stoyanov}]{conneau-etal-2020-unsupervised}
Alexis Conneau, Kartikay Khandelwal, Naman Goyal, Vishrav Chaudhary, Guillaume Wenzek, Francisco Guzm{\'a}n, Edouard Grave, Myle Ott, Luke Zettlemoyer, and Veselin Stoyanov. 2020.
\newblock \href {https://aclanthology.org/2020.acl-main.747} {Unsupervised cross-lingual representation learning at scale}.
\newblock In \emph{Proceedings of the 58th Annual Meeting of the Association for Computational Linguistics}, pages 8440--8451. Association for Computational Linguistics.

\bibitem[{Conneau et~al.(2018)Conneau, Rinott, Lample, Williams, Bowman, Schwenk, and Stoyanov}]{conneau-etal-2018-xnli}
Alexis Conneau, Ruty Rinott, Guillaume Lample, Adina Williams, Samuel Bowman, Holger Schwenk, and Veselin Stoyanov. 2018.
\newblock \href {https://doi.org/10.18653/v1/D18-1269} {{XNLI}: Evaluating cross-lingual sentence representations}.
\newblock In \emph{Proceedings of the 2018 Conference on Empirical Methods in Natural Language Processing}, pages 2475--2485, Brussels, Belgium. Association for Computational Linguistics.

\bibitem[{Costa-juss{\`a} et~al.(2017)Costa-juss{\`a}, Escolano, and Fonollosa}]{costa-jussa-etal-2017-byte}
Marta~R. Costa-juss{\`a}, Carlos Escolano, and Jos{\'e} A.~R. Fonollosa. 2017.
\newblock \href {https://doi.org/10.18653/v1/W17-4123} {Byte-based neural machine translation}.
\newblock In \emph{Proceedings of the First Workshop on Subword and Character Level Models in {NLP}}, pages 154--158, Copenhagen, Denmark. Association for Computational Linguistics.

\bibitem[{Costa-juss{à} et~al.(2022)Costa-juss{à}, Cross, Çelebi, Elbayad, Heafield, Heffernan, Kalbassi, Lam, Licht, Maillard, Sun, Wang, Wenzek, Youngblood, Akula, Barrault, Gonzalez, Hansanti, Hoffman, Jarrett, Sadagopan, Rowe, Spruit, Tran, Andrews, Ayan, Bhosale, Edunov, Fan, Gao, Goswami, Guzmán, Koehn, Mourachko, Ropers, Saleem, Schwenk, and Wang}]{costa2022no}
Marta~R. Costa-juss{à}, James Cross, Onur Çelebi, Maha Elbayad, Kenneth Heafield, Kevin Heffernan, Elahe Kalbassi, Janice Lam, Daniel Licht, Jean Maillard, Anna Sun, Skyler Wang, Guillaume Wenzek, Al~Youngblood, Bapi Akula, Loic Barrault, Gabriel~Mejia Gonzalez, Prangthip Hansanti, John Hoffman, Semarley Jarrett, Kaushik~Ram Sadagopan, Dirk Rowe, Shannon Spruit, Chau Tran, Pierre Andrews, Necip~Fazil Ayan, Shruti Bhosale, Sergey Edunov, Angela Fan, Cynthia Gao, Vedanuj Goswami, Francisco Guzmán, Philipp Koehn, Alexandre Mourachko, Christophe Ropers, Safiyyah Saleem, Holger Schwenk, and Jeff Wang. 2022.
\newblock \href {https://arxiv.org/abs/2207.04672} {No language left behind: Scaling human-centered machine translation}.
\newblock \emph{arXiv}.

\bibitem[{Daniels(1990)}]{daniels1990fundamentals}
Peter~T Daniels. 1990.
\newblock Fundamentals of grammatology.
\newblock \emph{Journal of the American Oriental Society}, pages 727--731.

\bibitem[{Davis(2012)}]{unicode-davis-2012}
Mark Davis. 2012.
\newblock \href {https://googleblog.blogspot.com/2012/02/unicode-over-60-percent-of-web.html} {Unicode over 60 percent of the web}.
\newblock Google Blog.

\bibitem[{Ding et~al.(2004)Ding, Peng, and Taft}]{ding2004nature}
Guosheng Ding, Danling Peng, and Marcus Taft. 2004.
\newblock The nature of the mental representation of radicals in {C}hinese: A priming study.
\newblock \emph{Journal of Experimental Psychology: Learning, Memory, and Cognition}, 30(2):530.

\bibitem[{e{B}ible(2023)}]{eBible}
e{B}ible. 2023.
\newblock \href {https://ebible.org/find/} {e{B}ible}.

\bibitem[{Guo et~al.(2020)Guo, Dai, Vrande{\v{c}}i{\'c}, and Al-Rfou}]{guo-etal-2020-wiki}
Mandy Guo, Zihang Dai, Denny Vrande{\v{c}}i{\'c}, and Rami Al-Rfou. 2020.
\newblock \href {https://aclanthology.org/2020.lrec-1.297} {{W}iki-40{B}: Multilingual language model dataset}.
\newblock In \emph{Proceedings of the Twelfth Language Resources and Evaluation Conference}, pages 2440--2452, Marseille, France. European Language Resources Association.

\bibitem[{Lin et~al.(2022)Lin, Mihaylov, Artetxe, Wang, Chen, Simig, Ott, Goyal, Bhosale, Du, Pasunuru, Shleifer, Koura, Chaudhary, O{'}Horo, Wang, Zettlemoyer, Kozareva, Diab, Stoyanov, and Li}]{lin-etal-2022-shot}
Xi~Victoria Lin, Todor Mihaylov, Mikel Artetxe, Tianlu Wang, Shuohui Chen, Daniel Simig, Myle Ott, Naman Goyal, Shruti Bhosale, Jingfei Du, Ramakanth Pasunuru, Sam Shleifer, Punit~Singh Koura, Vishrav Chaudhary, Brian O{'}Horo, Jeff Wang, Luke Zettlemoyer, Zornitsa Kozareva, Mona Diab, Veselin Stoyanov, and Xian Li. 2022.
\newblock \href {https://doi.org/10.18653/v1/2022.emnlp-main.616} {Few-shot learning with multilingual generative language models}.
\newblock In \emph{Proceedings of the 2022 Conference on Empirical Methods in Natural Language Processing}, pages 9019--9052, Abu Dhabi, United Arab Emirates. Association for Computational Linguistics.

\bibitem[{Moran et~al.(2014)Moran, McCloy, and Wright}]{moran2014phoible}
Steven Moran, Daniel McCloy, and Richard Wright. 2014.
\newblock \href {https://phoible.org/} {{PHOIBLE} online}.

\bibitem[{Moravcsik(2012)}]{moravcsik2012introducing}
Edith~A Moravcsik. 2012.
\newblock \emph{Introducing language typology}.
\newblock Cambridge University Press.

\bibitem[{Muennighoff et~al.(2023)Muennighoff, Wang, Sutawika, Roberts, Biderman, Le~Scao, Bari, Shen, Yong, Schoelkopf, Tang, Radev, Aji, Almubarak, Albanie, Alyafeai, Webson, Raff, and Raffel}]{muennighoff2022crosslingual}
Niklas Muennighoff, Thomas Wang, Lintang Sutawika, Adam Roberts, Stella Biderman, Teven Le~Scao, M~Saiful Bari, Sheng Shen, Zheng~Xin Yong, Hailey Schoelkopf, Xiangru Tang, Dragomir Radev, Alham~Fikri Aji, Khalid Almubarak, Samuel Albanie, Zaid Alyafeai, Albert Webson, Edward Raff, and Colin Raffel. 2023.
\newblock \href {https://doi.org/10.18653/v1/2023.acl-long.891} {Crosslingual generalization through multitask finetuning}.
\newblock In \emph{Proceedings of the 61st Annual Meeting of the Association for Computational Linguistics (Volume 1: Long Papers)}, pages 15991--16111, Toronto, Canada. Association for Computational Linguistics.

\bibitem[{Nettle(1995)}]{nettle1995segmental}
Daniel Nettle. 1995.
\newblock Segmental inventory size, word length, and communicative efficiency.
\newblock \emph{Linguistics}, 33(2):359--367.

\bibitem[{Perfetti and Liu(2005)}]{perfetti2005orthography}
Charles~A Perfetti and Ying Liu. 2005.
\newblock Orthography to phonology and meaning: Comparisons across and within writing systems.
\newblock \emph{Reading and Writing}, 18:193--210.

\bibitem[{Petrov et~al.(2024)Petrov, La~Malfa, Torr, and Bibi}]{petrov2024language}
Aleksandar Petrov, Emanuele La~Malfa, Philip Torr, and Adel Bibi. 2024.
\newblock \href {https://arxiv.org/abs/2305.15425} {Language model tokenizers introduce unfairness between languages}.
\newblock \emph{Advances in Neural Information Processing Systems}, 36.

\bibitem[{Ponti et~al.(2020)Ponti, Glava{\v{s}}, Majewska, Liu, Vuli{\'c}, and Korhonen}]{ponti-etal-2020-xcopa}
Edoardo~Maria Ponti, Goran Glava{\v{s}}, Olga Majewska, Qianchu Liu, Ivan Vuli{\'c}, and Anna Korhonen. 2020.
\newblock \href {https://doi.org/10.18653/v1/2020.emnlp-main.185} {{XCOPA}: A multilingual dataset for causal commonsense reasoning}.
\newblock In \emph{Proceedings of the 2020 Conference on Empirical Methods in Natural Language Processing (EMNLP)}, pages 2362--2376, Online. Association for Computational Linguistics.

\bibitem[{Scao et~al.(2022)Scao, Fan, Akiki, Pavlick, Ili'c, Hesslow, Castagn'e, Luccioni, Yvon, Gall{\'e}, Tow, Rush, Biderman, Webson, Ammanamanchi, Wang, Sagot, Muennighoff, del Moral, Ruwase et~al.}]{workshop2022bloom}
Teven~Le Scao, Angela Fan, Christopher Akiki, Elizabeth-Jane Pavlick, Suzana Ili'c, Daniel Hesslow, Roman Castagn'e, Alexandra~Sasha Luccioni, Franccois Yvon, Matthias Gall{\'e}, Jonathan Tow, Alexander~M. Rush, Stella~Rose Biderman, Albert Webson, Pawan~Sasanka Ammanamanchi, Thomas Wang, Beno{\^i}t Sagot, Niklas Muennighoff, Albert~Villanova del Moral, Olatunji Ruwase, et~al. 2022.
\newblock \href {https://arxiv.org/pdf/2211.05100.pdf} {{BLOOM}: A 176b-parameter open-access multilingual language model}.
\newblock \emph{arXiv}.

\bibitem[{{Unicode Consortium}(2022)}]{unicode-consortium}
{Unicode Consortium}. 2022.
\newblock \href {https://www.unicode.org/versions/Unicode15.0.0/UnicodeStandard-15.0.pdf} {{The Unicode Standard}}.

\bibitem[{van Esch et~al.(2022)van Esch, Lucassen, Ruder, Caswell, and Rivera}]{van-esch-etal-2022-writing}
Daan van Esch, Tamar Lucassen, Sebastian Ruder, Isaac Caswell, and Clara Rivera. 2022.
\newblock \href {https://aclanthology.org/2022.lrec-1.538} {Writing system and speaker metadata for 2,800+ language varieties}.
\newblock In \emph{Proceedings of the Thirteenth Language Resources and Evaluation Conference}, pages 5035--5046, Marseille, France. European Language Resources Association.

\bibitem[{Vatanen et~al.(2010)Vatanen, V{\"{a}}yrynen, and Virpioja}]{Vatanen10LREC}
Tommi Vatanen, Jaakko~J. V{\"{a}}yrynen, and Sami Virpioja. 2010.
\newblock Language identification of short text segments with n-gram models.
\newblock In \emph{Proceedings of the Seventh conference on International Language Resources and Evaluation (LREC'10)}, pages 3423--3430. European Language Resources Association (ELRA).

\bibitem[{Wagenmakers and Farrell(2004)}]{wagenmakers2004aic}
Eric-Jan Wagenmakers and Simon Farrell. 2004.
\newblock \href {https://link.springer.com/content/pdf/10.3758/BF03206482.pdf} {{AIC} model selection using {A}kaike weights}.
\newblock \emph{Psychonomic bulletin \& review}, 11:192--196.

\bibitem[{Williams and Bever(2010)}]{williams2010chinese}
Clay Williams and Thomas Bever. 2010.
\newblock Chinese character decoding: a semantic bias?
\newblock \emph{Reading and Writing}, 23:589--605.

\bibitem[{Xue et~al.(2022)Xue, Barua, Constant, Al-Rfou, Narang, Kale, Roberts, and Raffel}]{xue-etal-2022-byt5}
Linting Xue, Aditya Barua, Noah Constant, Rami Al-Rfou, Sharan Narang, Mihir Kale, Adam Roberts, and Colin Raffel. 2022.
\newblock \href {https://doi.org/10.1162/tacl_a_00461} {{B}y{T}5: Towards a token-free future with pre-trained byte-to-byte models}.
\newblock \emph{Transactions of the Association for Computational Linguistics}, 10:291--306.

\bibitem[{Zhang and Xu(2022)}]{zhang-xu-2022-byte}
Mengjiao Zhang and Jia Xu. 2022.
\newblock \href {https://aclanthology.org/2022.coling-1.388} {Byte-based multilingual {NMT} for endangered languages}.
\newblock In \emph{Proceedings of the 29th International Conference on Computational Linguistics}, pages 4407--4417, Gyeongju, Republic of Korea. International Committee on Computational Linguistics.

\end{thebibliography}

\appendix
\renewcommand\thesection{\Alph{section}}
\renewcommand\thefigure{\thesection.\arabic{figure}}
\renewcommand\thetable{\thesection.\arabic{table}}

\counterwithin{figure}{section}
\counterwithin{table}{section}
\section*{Appendices}

\section{NLLB Byte Premiums}\label{app:nllb_byte_premium_table}

Byte premiums calculated from NLLB are reported in Table \ref{tab:nllb_byte_premiums}.

\section{Byte Premiums After Compression}
\label{app:compression}
\begin{figure}[ht!]
\centering
\includegraphics[width=7.75cm]{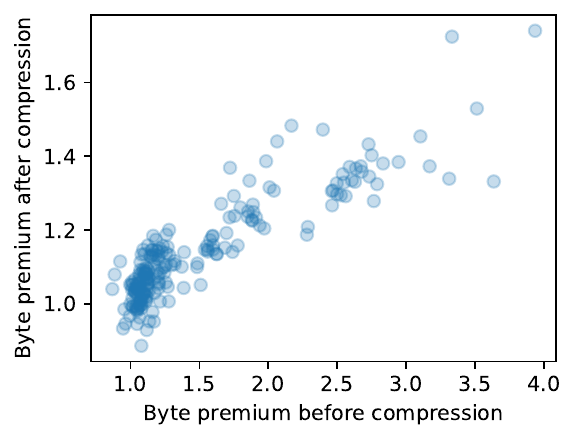}
\caption{Byte premiums before and after compression by $\texttt{gzip}$. Each point is a language's byte premium relative to English.}
\label{fig:after-compression}
\end{figure}

Byte premiums after compression by $\texttt{gzip}$, compared to those before compression, are plotted in Figure \ref{fig:after-compression}.

\setlength{\belowcaptionskip}{-0.3cm}
\setlength{\belowcaptionskip}{-0.0cm}

\section{Writing System Types} \label{app:writing_systems}
Our regressions in \S\ref{sec:estimating} require the script type for each language. The four possible script types are described below.

\paragraph{Alphabet} Alphabets are writing systems where each segment (either consonant or vowel) is represented by a symbol \citep{daniels1990fundamentals}. Latin script is one of the most widely used alphabets. Other alphabets include Greek, Cyrillic, and  Mkhedruli (Georgian). 

\paragraph{Abjad} Abjads are writing systems which represent each consonant with a symbol \citep{daniels1990fundamentals}, but vowels are often not represented. Arabic and Hebrew are written with abjads, for example.

\paragraph{Abugida} Abugidas, also sometimes referred to as \textit{neosyllabaries}, represent consonant-vowel sequences, often with vowel notation secondary to consonant notation \citep{daniels1990fundamentals}. Examples of abugidas include Devanagari (e.g. Hindi), Ge'ez (e.g. Amharic), and Canadian syllabics (e.g. Ojibwe).

\paragraph{Logography} Logographies are different from alphabets, abjads, and abugidas in that they represent semantic information as well as phonetic information. Chinese characters are the only logography that remains in use. The majority of Chinese characters are composed of one semantic component and one phonetic component \citep{williams2010chinese}. A relatively small number of characters are also pictographs or ideographs, representing only semantic information \citep{ding2004nature}.

\section{Validation from Different Datasets} \label{app:val_diff_datasets}

In Table \ref{tab:comparing_datasets_additional}, we report validation RMSEs for each regression (\S\ref{sec:evaluation}) when computing character entropies and bytes-per-character ratios from different datasets. Within each dataset, we separate the languages for which there are less than five other languages with the same script in the dataset from those which have five or more languages with the same script in the dataset. RMSE results are similar regardless of the dataset used to compute character entropies and bytes-per-character ratios.

\begin{table}[ht!]
\centering
\footnotesize
\begin{tabular}{|p{1.3cm}|p{1.85cm}|ccc|}
\cline{3-5}
\multicolumn{2}{c|}{} & \multicolumn{3}{c|}{Regression} \\
    \multicolumn{2}{c|}{} &
    \multicolumn{1}{c}{I} &
    \multicolumn{1}{c}{II} &
    \multicolumn{1}{c|}{III} \\
    \hline
    \multirow{2}{*}{NLLB} & Script ct. $\geq5$ & 0.201 & 0.244 & 0.240  \\ 
    \cline{2-1}
    & Script ct. $<5$ &0.700 & 0.744  & 0.637  \\
    \hline
    Flores & Script ct. $\geq5$ & 0.203 & 0.246 & 0.250 \\ 
    \cline{2-1}
    (20 lines) & Script ct. $<5$ &0.682 & 0.557  & 0.538  \\
    \hline
    Flores & Script ct. $\geq5$ & 0.204 & 0.252 & 0.254 \\ 
    \cline{2-1}
    (200) & Script ct. $<5$ & 0.702 & 0.615 & 0.544  \\
    \hline
    Flores & Script ct. $\geq5$ & 0.206 & 0.266 & 0.271 \\ 
    \cline{2-1}
    (2000) & Script ct. $<5$ & 0.703& 0.647 & 0.558 \\
    \hline
    Bible & Script ct. $\geq5$ & 0.272 & 0.294 & 0.298 \\ 
    \cline{2-1}
    (4 books) & Script ct. $<5$ & 0.766 & 0.680 & 0.577  \\
    \hline
    Bible & Script ct. $\geq5$ & 0.271 & 0.293 & 0.297 \\ 
    \cline{2-1}
    (1 book) & Script ct. $<5$ & 0.760 & 0.672 & 0.566  \\
    \hline
\end{tabular}
\normalsize
\caption{RMSEs when predicting byte premiums using different datasets to compute character entropies and bytes-per-character ratios. Results are separated into common and uncommon scripts.}
\label{tab:comparing_datasets_additional}
\end{table}

\section{Downstream Performance} \label{app:downstream_performance}

To evaluate the impact of byte premiums on downstream performance, we compile reported training data proportions (measured based on bytes) per language for existing massively multilingual models.
We adjust each training data proportion by dividing the reported proportion by the byte premium for that language. After re-scaling to sum to $1.0$, this provides the estimated effective proportion of data for each language. If adjusted data proportions are indeed ``better'' estimates of effective data quantities, then we expect them to predict downstream task performance better than the original reported training data proportions. 

We evaluate ten models from three model families: XGLM \citep{lin-etal-2022-shot}, BLOOM \citep{workshop2022bloom}, and mT0 \citep{muennighoff2022crosslingual}. We compile results from XGLM 7.5B, four sizes of BLOOM (560M, 1.1B, 3B, 7.1B), and five sizes of mT0 (small, base, large, xl, xxl). We use benchmark scores from five multilingual benchmarks: XStoryCloze \citep{lin-etal-2022-shot}, XCOPA \citep{ponti-etal-2020-xcopa}, XNLI \citep{conneau-etal-2018-xnli}, Wikipedia next word prediction \citep{guo-etal-2020-wiki}, and XWinograd \citep{muennighoff2022crosslingual}. These benchmarks cover 22 languages: Arabic, Bulgarian, German, Greek, English, Estonian, French, Haitian Creole, Hindi, Indonesian, Italian, Japanese, Burmese, Portuguese, Russian, Spanish, Swahili, Telugu, Turkish, Urdu, Vietnamese, and Chinese (simplified and traditional). Benchmark scores are compiled from the Big Science evaluation results on Hugging Face.\footnote{\href{https://huggingface.co/datasets/bigscience/evaluation-results}{https://huggingface.co/datasets/bigscience/evaluation-results}}

We fit two linear mixed effects models. Each predicts the benchmark score for each language (all scores between $0.0$ and $1.0$) from the training data proportion for that language (either the original proportion or those scaled according to our byte premiums) as well as language family, with random intercepts for model and task.
We calculate the AICs of the two non-nested models, along with their relative log likelihoods \citep{wagenmakers2004aic}. While the the data proportions scaled by byte premiums better predict benchmark performance (lower AIC and higher log likelihood), it is not a significant difference ($p = 0.13$), using significance testing as in \citet{wagenmakers2004aic}.
This non-significance may be because there are many other factors that impact downstream performance apart from dataset size. A larger meta-analysis would lead to more reliable inferences.

\newpage
\renewcommand{\thetable}{A.\arabic{table}}
\begin{table*}[b!]
\centering
\begin{tabular}{|l|l|}
\hline
\textbf{Language} & \textbf{Byte premium} \\ \hline
ace\_latn         & 1.2419926                  \\ \hline
afr\_latn         & 1.0373004                  \\ \hline
aka\_latn         & 1.5750612                  \\ \hline
als\_latn         & 1.1673181                  \\ \hline
amh\_ethi         & 1.7210862                  \\ \hline
arb\_arab         & 1.4651134                  \\ \hline
asm\_beng         & 2.5264323                  \\ \hline
ast\_latn         & 1.7490516                  \\ \hline
awa\_deva         & 2.7014324                  \\ \hline
ayr\_latn         & 1.0976628                  \\ \hline
azb\_arab         & 1.4901878                  \\ \hline
azj\_latn         & 1.0761036                  \\ \hline
bak\_cyrl         & 2.2716371                  \\ \hline
bam\_latn         & 1.2569819                  \\ \hline
ban\_latn         & 1.2695671                  \\ \hline
bem\_latn         & 1.1553301                  \\ \hline
ben\_beng         & 2.4308225                  \\ \hline
bho\_deva         & 2.5153669                  \\ \hline
bod\_tibt         & 2.6040539                  \\ \hline
bug\_latn         & 1.2279017                  \\ \hline
bul\_cyrl         & 1.8123562                  \\ \hline
cat\_latn         & 1.0926706                  \\ \hline
ceb\_latn         & 1.1134194                  \\ \hline
ces\_latn         & 1.0358867                  \\ \hline
ckb\_arab         & 1.6521034                  \\ \hline
ckb\_arab         & 1.6521034                  \\ \hline
cym\_latn         & 1.0265667                  \\ \hline
dan\_latn         & 1.0211031                  \\ \hline
deu\_latn         & 1.0537171                  \\ \hline
dik\_latn         & 1.1239299                  \\ \hline
diq\_latn         & 0.9590188                  \\ \hline
dyu\_latn         & 1.1545521                  \\ \hline
dzo\_tibt         & 3.2736977                  \\ \hline
ell\_grek         & 1.9673049                  \\ \hline
ewe\_latn         & 1.0783440                  \\ \hline
fao\_latn         & 1.1557437                  \\ \hline
fij\_latn         & 1.2107666                  \\ \hline
fin\_latn         & 1.0589051                  \\ \hline
fon\_latn         & 1.5413204                  \\ \hline
fra\_latn         & 1.1742064                  \\ \hline
fur\_latn         & 1.0672371                  \\ \hline
fuv\_latn         & 1.1109194                  \\ \hline
gla\_latn         & 0.9934613                  \\ \hline
gle\_latn         & 1.9749562                  \\ \hline
glg\_latn         & 1.0590246                  \\ \hline
guj\_gujr         & 2.1627759                  \\ \hline
hau\_latn         & 1.1766293                  \\ \hline
heb\_hebr         & 1.3555346                  \\ \hline
hin\_deva         & 2.3701629                  \\ \hline
hrv\_latn         & 0.9897218                  \\ \hline
hun\_latn         & 1.0199851                  \\ \hline
\end{tabular}
% \vfill\null 
\begin{tabular}{|l|l|}
% \centering
\hline
\textbf{Language} & \textbf{Byte premium} \\ \hline
hye\_armn         & 1.7241548                  \\ \hline
ibo\_latn         & 1.3451287                  \\ \hline
ilo\_latn         & 1.0765437                  \\ \hline
ind\_latn         & 1.1788023                  \\ \hline
isl\_latn         & 1.1543925                  \\ \hline
ita\_latn         & 1.0669230                  \\ \hline
jav\_latn         & 1.1468920                  \\ \hline
jpn\_jpan         & 1.3220250                  \\ \hline
kab\_latn         & 1.0287174                  \\ \hline
kac\_latn         & 1.3451812                  \\ \hline
kam\_latn         & 1.2177037                  \\ \hline
kan\_knda         & 2.6420061                  \\ \hline
kas\_arab         & 1.7762307                  \\ \hline
kas\_deva         & 2.5259810                  \\ \hline
kat\_geor         & 4.3381046                  \\ \hline
kbp\_latn         & 1.4408085                  \\ \hline
kea\_latn         & 0.7821679                  \\ \hline
khk\_cyrl         & 1.8046135                  \\ \hline
khm\_khmr         & 3.9051643                  \\ \hline
kik\_latn         & 1.2930516                  \\ \hline
kin\_latn         & 1.1340740                  \\ \hline
kir\_cyrl         & 1.9635570                  \\ \hline
kmr\_latn         & 1.0351712                  \\ \hline
knc\_arab         & 2.5022926                  \\ \hline
knc\_latn         & 1.1769876                  \\ \hline
kor\_hang         & 1.2933602                  \\ \hline
lao\_laoo         & 2.7071355                  \\ \hline
lij\_latn         & 1.1438412                  \\ \hline
lin\_latn         & 1.1393024                  \\ \hline
lit\_latn         & 1.0300780                  \\ \hline
ltg\_latn         & 1.0028570                  \\ \hline
ltz\_latn         & 1.2253827                  \\ \hline
lug\_latn         & 1.2175185                  \\ \hline
luo\_latn         & 1.0358323                  \\ \hline
lus\_latn         & 1.1689564                  \\ \hline
lvs\_latn         & 1.2070388                  \\ \hline
mag\_deva         & 2.5555142                  \\ \hline
mai\_deva         & 2.3896953                  \\ \hline
mal\_mlym         & 2.8852389                  \\ \hline
mar\_deva         & 2.4793638                  \\ \hline
min\_latn         & 0.9497956                  \\ \hline
mkd\_cyrl         & 1.8349890                  \\ \hline
mlt\_latn         & 1.0884567                  \\ \hline
mni\_beng         & 3.0027416                  \\ \hline
mos\_latn         & 1.1413713                  \\ \hline
mri\_latn         & 1.1826053                  \\ \hline
mya\_mymr         & 5.1034592                  \\ \hline
nld\_latn         & 1.0516739                  \\ \hline
nob\_latn         & 0.9977426                  \\ \hline
npi\_deva         & 2.4202344                  \\ \hline
nus\_latn         & 1.2935254                  \\ \hline
\end{tabular}
\begin{tabular}{|l|l|}
% \centering
\hline
\textbf{Language} & \textbf{Byte premium} \\ \hline
oci\_latn         & 1.0146652                  \\ \hline
ory\_orya         & 2.5109372                  \\ \hline
pag\_latn         & 1.0439418                  \\ \hline
pan\_guru         & 2.2208951                  \\ \hline
pbt\_arab         & 1.7366557                  \\ \hline
pes\_arab         & 1.5973940                  \\ \hline
plt\_latn         & 1.1512264                  \\ \hline
pol\_latn         & 1.0774161                  \\ \hline
por\_latn         & 1.0979270                  \\ \hline
quy\_latn         & 1.1639224                  \\ \hline
ron\_latn         & 1.1151666                  \\ \hline
run\_latn         & 1.1193204                  \\ \hline
rus\_cyrl         & 1.8228284                  \\ \hline
sag\_latn         & 1.1632489                  \\ \hline
san\_deva         & 2.5428913                  \\ \hline
sat\_beng         & 2.1131754                  \\ \hline
shn\_mymr         & 2.8224643                  \\ \hline
sin\_sinh         & 2.4463506                  \\ \hline
slk\_latn         & 1.0415468                  \\ \hline
slv\_latn         & 0.9722273                  \\ \hline
sna\_latn         & 1.1192729                  \\ \hline
snd\_arab         & 1.5880165                  \\ \hline
som\_latn         & 1.4224149                  \\ \hline
sot\_latn         & 1.1661078                  \\ \hline
spa\_latn         & 1.0838621                  \\ \hline
srp\_cyrl         & 1.4249495                  \\ \hline
sun\_latn         & 1.0970417                  \\ \hline
swe\_latn         & 1.0210256                  \\ \hline
swh\_latn         & 1.0696621                  \\ \hline
tam\_taml         & 2.7292892                  \\ \hline
taq\_latn         & 1.2093634                  \\ \hline
tat\_cyrl         & 1.8543562                  \\ \hline
tel\_telu         & 2.6198705                  \\ \hline
tgk\_cyrl         & 1.7469201                  \\ \hline
tgl\_latn         & 1.1176348                  \\ \hline
tir\_ethi         & 1.7631466                  \\ \hline
tuk\_latn         & 1.7850561                  \\ \hline
tur\_latn         & 1.0444815                  \\ \hline
tzm\_tfng         & 1.9259158                  \\ \hline
uig\_arab         & 2.3082357                  \\ \hline
ukr\_cyrl         & 1.7514786                  \\ \hline
umb\_latn         & 1.1673612                  \\ \hline
urd\_arab         & 1.7079714                  \\ \hline
uzn\_latn         & 1.6455453                  \\ \hline
vie\_latn         & 1.3493725                  \\ \hline
wol\_latn         & 1.0787309                  \\ \hline
xho\_latn         & 1.1988860                  \\ \hline
ydd\_hebr         & 1.8074376                  \\ \hline
yor\_latn         & 1.3750599                  \\ \hline
zsm\_latn         & 1.1438457                  \\ \hline
zul\_latn         & 1.1639372                  \\ \hline
\end{tabular}
\caption{NLLB byte premiums. The byte premium for eng\_latn is $1.0$. Each language code is comprised of the ISO 639-3 (language) and ISO 15924 (script) codes separated by an underscore.}
\label{tab:nllb_byte_premiums}
\end{table*}

\end{document}